\documentclass[conference]{IEEEtran}
\IEEEoverridecommandlockouts

\usepackage{cite}
\usepackage{amsmath,amssymb,amsfonts}
\usepackage{algorithmic}
\usepackage{graphicx}
\usepackage{textcomp}
\usepackage{xcolor}
\usepackage{enumitem}
\usepackage{graphicx}
\usepackage[caption=false,font=footnotesize]{subfig}
\usepackage[font=scriptsize]{caption}
\usepackage{booktabs}
\usepackage{cite}
\usepackage{amsmath,amssymb,amsfonts}
\usepackage{graphicx}
\usepackage{textcomp}
\usepackage{xcolor}
\usepackage{url}
\usepackage{multirow}
\usepackage{caption}
\usepackage[hidelinks]{hyperref}

\def\BibTeX{{\rm B\kern-.05em{\sc i\kern-.025em b}\kern-.08em
    T\kern-.1667em\lower.7ex\hbox{E}\kern-.125emX}}

\begin{document}


\title{Modeling Patient Care Trajectories with Transformer Hawkes Processes%
}

\author{
\IEEEauthorblockN{Saumya Pandey}
\IEEEauthorblockA{
\textit{Computational and Data-Enabled Sciences} \\
\textit{University at Buffalo} \\
Buffalo, USA \\
spandey5@buffalo.edu
}
\and
\IEEEauthorblockN{Varun Chandola}
\IEEEauthorblockA{
\textit{Department of Computer Science and Engineering} \\
\textit{University at Buffalo} \\
Buffalo, USA \\
chandola@buffalo.edu
}
}

\maketitle

\begin{abstract}
Patient healthcare utilization consists of irregularly time-stamped events, including outpatient visits, inpatient admissions, and emergency department encounters, which together form individualized care trajectories. Modeling such trajectories is essential for understanding utilization patterns and anticipating future care needs, but remains challenging due to temporal irregularity and severe class imbalance in real-world healthcare data. In this work, we build upon the Transformer Hawkes Process framework to model patient healthcare utilization trajectories in continuous time under realistic, highly imbalanced data conditions. By combining Transformer-based history encoding with a Hawkes point process formulation, the framework captures event-driven temporal dynamics and supports joint prediction of the next event type and the time gap until its occurrence. While prior Transformer-based point process models offer expressive temporal representations, they have not been systematically evaluated under the extreme event-type imbalance characteristic of healthcare utilization data. To address this limitation, we incorporate an imbalance-aware optimization strategy based on weighted cross-entropy, with class weights defined by the inverse square root of event frequencies. This approach preserves the natural temporal structure of patient trajectories while substantially improving sensitivity to rare but clinically significant inpatient and emergency events, without resorting to resampling or altering the underlying event distribution. We evaluate the resulting framework on real-world patient utilization data and demonstrate that this imbalance-aware extension enables more effective modeling of highly skewed care trajectories and improves both event-type prediction and time-to-event estimation. The resulting predictions provide clinically meaningful insights that can support healthcare systems in identifying, monitoring, and managing high-need, high-cost patient populations.
\end{abstract}


\section{Introduction}

Healthcare utilization reflects the sequence of interactions that patients have with the healthcare system over time, including outpatient visits, inpatient admissions, and emergency department encounters. Prior work has shown that these interactions form distinct longitudinal utilization trajectories across care settings~\cite{article}. These interactions are not isolated events; rather, they form temporally ordered care trajectories that capture the evolution of a patient’s health status, care needs, and engagement with the healthcare system. Understanding and modeling such trajectories is critical for a wide range of healthcare objectives, including early identification of risk~\cite{10.1001/jamanetworkopen.2018.4273}, care coordination~\cite{10.1001/jamanetworkopen.2018.4273}, resource planning~\cite{10.1001/jamanetworkopen.2018.4273}, and cost containment~\cite{10.1001/jamanetworkopen.2018.4273}. This is particularly important for high-need, high-cost patient populations, as improved trajectory modeling can support more efficient and targeted care delivery~\cite{berwick2008tripleaim}.

From a data perspective, healthcare utilization trajectories present several inherent challenges. Events occur irregularly in continuous time, with substantial variability in both the frequency and spacing of visits across patients. Individual trajectories often exhibit long periods of routine outpatient care followed by sudden inpatient admissions or emergency department visits, reflecting acute changes in health status. These event sequences are highly heterogeneous across patients and within individual histories, making it difficult to apply modeling approaches that assume regular sampling or homogeneous temporal patterns~\cite{8086133}. As a result, capturing the underlying temporal dynamics of healthcare utilization remains a challenging and essential problem.

A further complication arises from the strong imbalance in event types observed in real-world healthcare data. Outpatient visits typically dominate patient histories, while inpatient admissions and emergency department encounters occur far less frequently. Despite their relative rarity, these events are often of greater clinical and operational significance, as they are associated with acute illness and elevated patient risk. Models that fail to account for this imbalance may achieve high overall predictive accuracy while systematically underperforming on rare yet clinically critical event types~\cite{rajkomar2018scalable}. Consequently, effective trajectory modeling requires not only temporal expressiveness but also robustness to highly skewed and imbalanced event distributions.

In this work, we model healthcare utilization trajectories directly in continuous time, avoiding discretization of event timelines. We leverage a Transformer-based Hawkes Process (THP) framework~\cite{zuo2021transformerhawkesprocess} to jointly model multi-type healthcare events and their timing through a conditional intensity formulation. While prior Transformer-based point process models primarily emphasize architectural expressiveness and flexible temporal dependency modeling~\cite{zuo2021transformerhawkesprocess}, they typically address event imbalance through dataset-specific preprocessing or heuristic strategies rather than explicitly incorporating imbalance-aware training objectives. This limitation is particularly pronounced in healthcare utilization settings, where outpatient visits dominate patient histories and clinically critical inpatient and emergency events are relatively rare.

To address this gap, we introduce an imbalance-aware training objective based on a weighted cross-entropy loss, with class weights defined using the inverse square root of event frequencies. Unlike resampling-based strategies, which may distort the temporal structure of patient trajectories and reduce clinical realism, the proposed loss formulation preserves the natural event sequence while improving sensitivity to rare but high-impact healthcare events. Through empirical evaluation on real-world patient utilization data, we demonstrate improved prediction of both event types and timing under severe class imbalance.

The key contributions of this paper are summarized as follows:
\begin{itemize}
\item We formulate patient healthcare utilization trajectories as a marked temporal point process and adapt a Transformer Hawkes Process framework to model irregular, multi-type healthcare events with long-range temporal dependencies.
\item We introduce an imbalance-aware training objective using inverse square-root frequency–weighted cross-entropy, improving sensitivity to rare but clinically critical inpatient and emergency events without altering the underlying event distribution.
\item We evaluate the proposed approach on real-world healthcare utilization data and demonstrate improved modeling of individualized, imbalanced care trajectories and prediction of future healthcare events and timing.
\item We provide qualitative interpretability analyses illustrating how the model captures clinically intuitive short-term risk and long-term patient vulnerability across heterogeneous care trajectories.
\end{itemize}

\section{Related Work}

\subsection{Healthcare Trajectory Modeling}

Modeling patient trajectories from electronic health records (EHRs) has been an active area of research~\cite{8086133,pham2017deepcaredeepdynamicmemory,rajkomar2018scalable}, motivated by applications such as readmission prediction~\cite{ASHFAQ2019103256}, risk stratification~\cite{Deeppatient}, and disease progression modeling~\cite{9502579}. Early approaches typically rely on supervised machine learning over aggregated or snapshot-based features extracted from structured EHR data, including demographic, diagnostic, procedural, and nursing variables~\cite{Deeppatient}. Such models, often based on generalized linear models or tree-based ensembles, have demonstrated utility for specific prediction tasks but operate on static representations of patient history and do not explicitly capture longitudinal care dynamics~\cite{info:doi/10.2196/56671}.

To better model temporal dependencies, sequential deep learning approaches, for instance, recurrent neural networks (RNNs) and long short-term memory (LSTM) architectures have been widely adopted in healthcare applications~\cite{ASHFAQ2019103256,pham2017deepcaredeepdynamicmemory}. These models treat patient records as ordered sequences of visits and capture temporal evolution implicitly through recurrent hidden states. While effective for short-term prediction tasks, such representations can obscure irregular temporal spacing between events and introduce arbitrary temporal boundaries, limiting their ability to capture long-range temporal dependencies in heterogeneous patient trajectories.

\subsection{Temporal Point Processes in Healthcare}

Temporal point processes provide a principled framework for modeling event-driven healthcare data in continuous time, representing patient histories as sequences of time-stamped events governed by a conditional intensity function~\cite{xu2019tpp}. Hawkes processes~\cite{10.1093/biomet/58.1.83} have been applied to healthcare data to model disease progression~\cite{9502579}, hospital admissions~\cite{10.1002/sam.11409}, and care utilization patterns~\cite{10.1002/sam.11409}. However, classical Hawkes models rely on fixed parametric triggering kernels, which may limit expressiveness when modeling complex, multi-type, and long-range temporal dependencies~\cite{mei2017neuralhawkesprocessneurally}.

Neural temporal point process models~\cite{mei2017neuralhawkesprocessneurally} and Transformer-based Hawkes processes~\cite{zuo2021transformerhawkesprocess} address these limitations by introducing data-driven representations of event dependencies. While these models significantly enhance architectural expressiveness, they primarily focus on temporal modeling capacity and do not explicitly address severe event-type imbalance commonly observed in healthcare utilization data.

\subsection{Interpretability in Healthcare Machine Learning}

Interpretability is widely recognized as a key requirement for the adoption of machine learning models in clinical settings~\cite{pmlr-v106-tonekaboni19a}. Prior work emphasizes that clinicians’ trust in predictive systems depends on the ability to understand, contextualize, and rationalize model outputs~\cite{pmlr-v106-tonekaboni19a}. Attention-based models~\cite{vaswani2023attentionneed} and intensity-based temporal point process formulations~\cite{10.1093/biomet/58.1.83} naturally support interpretability through mechanisms such as attention weights, temporal attribution, and conditional intensity functions. Despite this, the integration of interpretable temporal representations with learning strategies that remain robust under severe class imbalance has received limited attention, motivating the interpretability analyses presented in this work~\cite{Bharati_2024}.

\section{Background}

\subsection{GLM Baseline}

A generalized linear model (GLM) is a class of statistical models that relates input features to an outcome through a linear predictor and a suitable link function, allowing flexible modeling of different data distributions. Formally, a GLM assumes:
\begin{equation}
g(\mathbb{E}[y \mid \mathbf{x}]) = \beta_0 + \boldsymbol{\beta}^\top \mathbf{x}
\end{equation}
where $g(\cdot)$ is the link function, $\mathbf{x}$ denotes input features, and $\boldsymbol{\beta}$ are learnable parameters.

We implement a GLM-based baseline by summarizing each patient’s recent history using a sliding time window of 100 days, capturing counts of prior visit types (IP, OP, ED) along with temporal features such as time since last visit and time since trajectory start. The next-event type is modeled using multinomial logistic regression:
\begin{equation}
P(y = k \mid \mathbf{x}) = \frac{\exp(\boldsymbol{\beta}_k^\top \mathbf{x})}{\sum_{j} \exp(\boldsymbol{\beta}_j^\top \mathbf{x})}
\end{equation}
while the time-to-next-event is modeled using a Tweedie (Gamma) regression with a log link:
\begin{equation}
\mathbb{E}[\Delta t \mid \mathbf{x}] = \exp(\boldsymbol{\beta}^\top \mathbf{x})
\end{equation}
ensuring positive and skew-aware predictions.

This baseline captures short-term temporal signals through aggregated features but does not explicitly model sequential dependencies or continuous-time event interactions, motivating the use of more expressive models such as Transformer Hawkes Processes.

\subsection{Temporal Point Processes}

Temporal point processes (TPPs)~\cite{daley2003introduction,rasmussen2018lecturenotestemporalpoint} provide a principled mathematical framework for modeling sequences of discrete events that occur irregularly over continuous time. Unlike classical time-series models, which assume observations at fixed and regularly spaced intervals, TPPs represent each event explicitly by its occurrence time and, optionally, an associated mark encoding event-specific information such as type or category. This formulation is particularly well suited for healthcare utilization data, where events such as outpatient visits, inpatient admissions, and emergency department encounters occur asynchronously and exhibit substantial variability in both timing and frequency across patients.

A central advantage of TPPs lies in their ability to preserve fine-grained temporal information. In healthcare settings, the time elapsed between events often carries clinically meaningful signals—for example, rapid readmission following discharge may indicate unresolved clinical issues, while extended gaps between visits may reflect stable disease management or successful outpatient care. By operating directly in continuous time, TPPs avoid the need for arbitrary discretization choices that can obscure such temporal patterns and distort underlying dynamics.

Formally, a temporal point process is characterized by a conditional intensity function, which defines the instantaneous rate at which events are expected to occur given the history of past events. This intensity function serves as the primary object of modeling and inference, governing both the likelihood of future events and their expected timing.

A temporal point process can be represented as a sequence of event times $\{\tau_j\}_{j \in J}$, where each $\tau_j$ denotes the occurrence time of an event. The conditional intensity function $\lambda(t)$ is defined as:

\begin{equation}
\lambda(t) = \lim_{\Delta t \to 0} \frac{\mathbb{P}(N(t+\Delta t) - N(t) = 1 \mid \mathcal{H}_t)}{\Delta t}
\end{equation}

where $N(t)$ denotes the number of events that have occurred up to time $t$, and $\mathcal{H}_t$ represents the history of all past events prior to time $t$. Intuitively, $\lambda(t)$ quantifies the instantaneous rate at which an event is expected to occur at time $t$, conditioned on the observed history \cite{daley2003introduction}.

\subsection{Hawkes Processes}
Hawkes processes~\cite{10.1093/biomet/58.1.83} are a class of self-exciting temporal point processes in which the occurrence of past events increases the likelihood of future events within a temporal neighborhood. This behavior is modeled through a history-dependent conditional intensity function defined as:

\begin{equation}
\lambda(t) = \mu + \sum_{t_i < t} \alpha \, g(t - t_i)
\end{equation}

where $\mu$ is the baseline intensity, $\alpha$ controls the influence of past events, and $g(t - t_i)$ is a kernel function that models the temporal decay of influence from a past event occurring at time $t_i$. This formulation captures the clustering effect commonly observed in real-world event sequences, where events tend to occur in bursts rather than independently~\cite{bacry2015hawkes}.

In classical Hawkes formulations, the influence of past events is governed by predefined parametric triggering kernels, such as exponential or power-law decay functions. These kernels specify how excitation effects diminish over time and provide interpretability, as their parameters directly encode the magnitude and temporal decay of influence. Such properties have made Hawkes processes attractive for modeling event-driven phenomena across diverse domains, including seismology~\cite{2023SpaSt..5400728K}, finance~\cite{bacry2015hawkes}, social interactions~\cite{Mitchell_2009,salehi2019learninghawkesprocesseshandful}, and epidemiology~\cite{salehi2019learninghawkesprocesseshandful,10.1098/rstb.2016.0308}.

However, when applied to patient-level healthcare data, classical Hawkes processes face notable challenges. Healthcare utilization trajectories are inherently multi-type, highly heterogeneous, and often dominated by a small number of frequent event types alongside rare but clinically critical events. Moreover, temporal dependencies in healthcare data may be nonlinear, delayed, patient-specific, and span long time horizons, making them difficult to capture using fixed parametric kernel structures. These limitations motivate the development of more flexible modeling approaches that retain the continuous-time foundation of Hawkes processes while relaxing restrictive assumptions on temporal dynamics.

\subsection{Neural and Transformer-Based Hawkes Models}
To overcome the limitations of parametric triggering kernels, neural temporal point process models have been proposed~\cite{mei2017neuralhawkesprocessneurally}. In these approaches, neural networks are used to encode the history of past events and parameterize the conditional intensity function, replacing fixed functional forms with data-driven representations learned directly from data. This enables the model to capture complex temporal dependencies, including nonlinear interactions, inhibition effects, and multiple temporal scales, while preserving a likelihood-based training objective.

Transformer-based Hawkes Process models~\cite{zuo2021transformerhawkesprocess} further extend this paradigm by leveraging self-attention mechanisms to encode historical events. Self-attention allows the model to dynamically weight the relevance of past events when forming predictions, independent of their position in the sequence or their temporal distance. This property is particularly advantageous in healthcare utilization modeling, where clinically meaningful events may be separated by long and irregular time gaps, and both recent and distant history can influence future risk.

Importantly, Transformer-based Hawkes models retain a probabilistic and continuous-time foundation through their use of conditional intensity functions. As a result, they combine the interpretability and theoretical grounding of point process models with the representational flexibility of attention-based sequence encoders. This hybrid formulation enables principled modeling of irregular, multi-type event sequences while supporting interpretability through mechanisms such as attention weights, intensity trajectories, and temporal attribution over patient histories.

Together, neural and Transformer-based Hawkes processes provide a powerful modeling framework for complex healthcare utilization data, offering a balance between expressive temporal modeling, probabilistic rigor, and interpretability. These properties make them well suited for high-stakes clinical applications that require both accurate prediction and transparent temporal reasoning.

\section{Core Methodology}

\subsection{Problem Formulation}

Let a patient trajectory be represented as a sequence of historical events:
\begin{equation}
\mathcal{H}_i = \{(t_1, k_1), (t_2, k_2), \dots, (t_i, k_i)\},
\end{equation}
where \( t_j \in \mathbb{R}^{+} \) denotes the time of the \(j\)-th event, and \( k_j \in \{1, \dots, K\} \) denotes the corresponding event type. In this work, \(K = 3\), corresponding to inpatient (IP), outpatient (OP), and emergency department (ED) events. The history \( \mathcal{H}_i \) contains all events observed up to time \( t_i \).

The objective is to model the conditional distribution of the next event \((t_{i+1}, k_{i+1})\) given the observed history:
\begin{equation}
p(t_{i+1}, k_{i+1} \mid \mathcal{H}_i).
\end{equation}

\subsection{Methodology Overview}

We adopt the Transformer Hawkes Process (THP) framework to model patient healthcare utilization as a multivariate temporal point process in continuous time. The model represents each patient trajectory as a sequence of timestamped events with discrete event types corresponding to inpatient, outpatient, and emergency department encounters.

Patient history is encoded using a Transformer-based self-attention mechanism, which produces a latent representation of past events without relying on recurrent state updates. This representation is used to parameterize Hawkes process conditional intensity functions, allowing future event dynamics to depend explicitly on the entire observed history.

The model jointly estimates (i) the probability distribution over the next event type and (ii) the inter-event time until the next occurrence. Event timing is modeled directly in continuous time through the conditional intensity formulation, avoiding discretization of the temporal axis.

To accommodate the severe imbalance in event-type frequencies present in healthcare utilization data, we modify the standard THP training objective by incorporating an imbalance-aware weighted cross-entropy loss for event-type prediction. This adjustment affects optimization only and does not alter the underlying model architecture or intensity formulation.

\subsection{Transformer-Based History Encoding}

Each historical event is embedded using both event-type information and temporal information. The input representation for the \(j\)-th event is defined as:
\begin{equation}
\mathbf{x}_j = \mathbf{e}_{k_j} + \mathbf{e}_{\Delta t_j},
\end{equation}
where \( \mathbf{e}_{k_j} \) is a learnable embedding corresponding to the event type \(k_j\), and \( \Delta t_j = t_j - t_{j-1} \) denotes the inter-event time between consecutive events.

The temporal embedding \( \mathbf{e}_{\Delta t_j} \) encodes when an event occurred relative to the previous event and is obtained using continuous sinusoidal temporal encoding adapted from Transformer positional encodings. Inter-event times are log-scaled and normalized prior to temporal encoding to stabilize training. Specifically, for embedding dimension \(d\),
\begin{align}
\mathbf{e}_{\Delta t_j}^{(2m)} &= \sin\left( \frac{\Delta t_j}{10000^{2m/d}} \right), \\
\mathbf{e}_{\Delta t_j}^{(2m+1)} &= \cos\left( \frac{\Delta t_j}{10000^{2m/d}} \right),
\end{align}
where \( m = 0, \dots, \frac{d}{2} - 1 \).

The resulting event representations are passed through a Transformer encoder with causal masking, producing contextualized hidden representations that summarize the patient’s historical healthcare utilization.

\subsection{Conditional Intensity Modeling}

After temporal and event-type encoding, the sequence of event representations is passed through a Transformer encoder with causal masking to obtain contextualized hidden states. Let $\mathbf{h}_i$ denote the hidden representation corresponding to the patient history observed up to time $t_i$. This representation serves as a compact summary of the patient’s historical healthcare utilization.

Building on this representation, we adopt a Hawkes process--based formulation, in which the occurrence of future events at time $t$ is governed by a conditional intensity function conditioned on the observed history.

\subsubsection{Event-Type--Specific Conditional Intensities}

Given the heterogeneous nature of healthcare utilization, we define separate conditional intensity functions for each event type. Specifically, for each event type $k \in \{\mathrm{IP}, \mathrm{OP}, \mathrm{ED}\}$, the conditional intensity function is parameterized as:
\begin{equation}
\lambda_k(t \mid \mathcal{H}_{i})
=
\mathrm{softplus}_{\beta}
\left(
\mathbf{w}_k^{\top}\mathbf{h}_{i}
+
b_k
+
\alpha_k \, s
\right),
\label{eq:intensity}
\end{equation}
where $\mathbf{w}_k$ and $b_k$ are learnable parameters associated with event type $k$, and $\mathbf{h}_i$ is the Transformer hidden state summarizing the observed history up to time $t_i$. Here, $b_k$ represents the baseline log-intensity for event type $k$, modeling its inherent risk in the absence of historical influence.
 The term $s \in [0,1]$ denotes a normalized within-interval time variable over the interval $(t_i, t_{i+1})$, and $\alpha_k$ is a learnable parameter controlling the temporal evolution of the intensity within the interval. The term $\alpha_k s$ provides a continuous interpolation of event-type--specific risk between consecutive observations, allowing the conditional intensity to evolve smoothly as time elapses since the most recent event.

The $\beta$-softplus transformation ensures that the intensity function remains strictly positive while improving numerical stability.

This formulation enables the model to capture distinct temporal risk profiles for inpatient admissions, outpatient visits, and emergency department encounters.

\subsubsection{Overall Intensity Function}

The total conditional intensity governing the occurrence of the next event of any type is obtained by summing the event-type--specific intensities:
\begin{equation}
\lambda(t \mid \mathcal{H}_{i})
=
\sum_{k}
\lambda_k(t \mid \mathcal{H}_{i}).
\label{eq:total_intensity}
\end{equation}

The overall intensity determines the timing of the next event, while the relative magnitudes of the individual event-type--specific intensities determine the event type.

\subsubsection{Event-Type Probability Estimation}

The probability of the next event being of type $k$, conditioned on the observed history $\mathcal{H}_{i}$, is derived by normalizing the event-type--specific intensities:
\begin{equation}
p(k \mid \mathcal{H}_{i})
=
\frac{\lambda_k(t \mid \mathcal{H}_{i})}
{\sum_{k'} \lambda_{k'}(t \mid \mathcal{H}_{i})}.
\label{eq:type_prob}
\end{equation}

This formulation ensures consistency between event-type prediction and the underlying temporal point process, enabling joint modeling of event type and event time.

\subsection{Likelihood Function and Training Objective}

The model is trained using a temporal point process likelihood formulation. Given a patient trajectory
$\mathcal{H} = \{(t_1, k_1), \ldots, (t_N, k_N)\}$, the log-likelihood of the observed event sequence is defined as:
\begin{equation}
\mathcal{L}_{\text{NLL}}
=
\sum_{j=2}^{N}
\log \lambda_{k_j}(t_j \mid \mathcal{H}_{j-1})
-
\int_{t_{j-1}}^{t_j}
\lambda(t \mid \mathcal{H}_{j-1}) \, dt,
\label{eq:loglik}
\end{equation}
where $\lambda_{k_j}(t_j \mid \mathcal{H}_{j-1})$ denotes the conditional intensity of the observed event type at time $t_j$, and
$\lambda(t \mid \mathcal{H}_{j-1}) = \sum_k \lambda_k(t \mid \mathcal{H}_{j-1})$ represents the total conditional intensity.

The first term encourages high intensity for observed events, while the second term penalizes excessive intensity during periods in which no events occur. The model is trained by minimizing the negative log-likelihood over all patient trajectories.

\subsection{Handling Class Imbalance}
Many real-world problems face the issue of imbalance in the frequency of different event types. This is typical in the healthcare setting like the one discussed here, with outpatient visits dominating the event distribution, followed by emergency encounters and relatively sparse inpatient admissions. To mitigate bias toward frequent event types, we incorporate a weighted event-type classification loss.

Specifically, we apply a weighted cross-entropy (CE) loss for event-type prediction, where each event type $k$ is assigned a weight inversely proportional to the square root of its empirical frequency:
\begin{equation}
w_k = \frac{1}{\sqrt{f_k}},
\end{equation}
where $f_k$ denotes the frequency of event type $k$ in the training data.

This weighting scheme encourages improved recognition of rare but clinically significant events, such as inpatient admissions and emergency department encounters, while maintaining stable optimization. The same weighting strategy is applied consistently across neural baselines to ensure a fair comparison.

\subsection{Prediction of Time-to-Event}

In addition to event-type prediction, the model estimates the time until the next clinical event occurs. Time-to-event prediction is derived implicitly from the learned conditional intensity function, which defines a probability distribution over future event times.

To accommodate the heavy-tailed nature of inter-event time gaps in healthcare utilization data, the model predicts the transformed time gap $\log(1 + \Delta t_j)$, where $\Delta t_j = t_j - t_{j-1}$. This transformation stabilizes learning by reducing the influence of extreme temporal gaps while preserving relative timing information.

At inference time, predicted values are transformed back to the original scale to obtain absolute time-to-event estimates, expressed in days. Model performance is evaluated using the median absolute error (MedAE), which provides a robust measure of typical temporal prediction accuracy and is less sensitive to rare but extreme delays than mean-based metrics.

\subsection{Overall Training Objective}

The final training objective combines the point-process negative log-likelihood with the weighted event-type classification loss:
\begin{equation}
\mathcal{L}_{\text{total}}
=
\mathcal{L}_{\text{NLL}}
+
\gamma \mathcal{L}_{\text{type}}
\end{equation}
where $\gamma$ controls the relative contribution of event-type supervision.

\begin{table*}[t]
\centering
\caption{Comparison of event-type prediction and time-to-event prediction performance across models.
Macro-F1 and event-type F1 scores highlight balanced performance under severe class imbalance.
Time prediction is evaluated using MedAE (days).}
\label{tab:merged_performance}
\resizebox{\textwidth}{!}{
\begin{tabular}{lccccc}
\toprule
\multirow{2}{*}{\textbf{Model}} 
& \multicolumn{4}{c}{\textbf{Event-Type Prediction}} 
& \textbf{Time-to-Event Prediction} \\
\cmidrule(lr){2-5} \cmidrule(lr){6-6}
& \textbf{Macro-F1} 
& \textbf{IP F1} 
& \textbf{ED F1}
& \textbf{OP F1}
& \textbf{MedAE (days)} \\
\midrule
GLM 
& 0.290 
& 0.0002 
& 0.0070 
& 0.9511
& 28.25 \\
LSTM Hawkes (weighted CE)
& 0.432 
& 0.108 
& 0.246 
& 0.943 
& 18.00 \\
Transformer Hawkes (THP, weighted CE)
& \textbf{0.480} 
& \textbf{0.162} 
& \textbf{0.334} 
& 0.942 
& \textbf{13.00} \\
\bottomrule
\end{tabular}
}
\end{table*}

\begin{figure*}[t]
\centering
\subfloat[Conditional intensity curves]{%
  \includegraphics[width=0.31\textwidth]{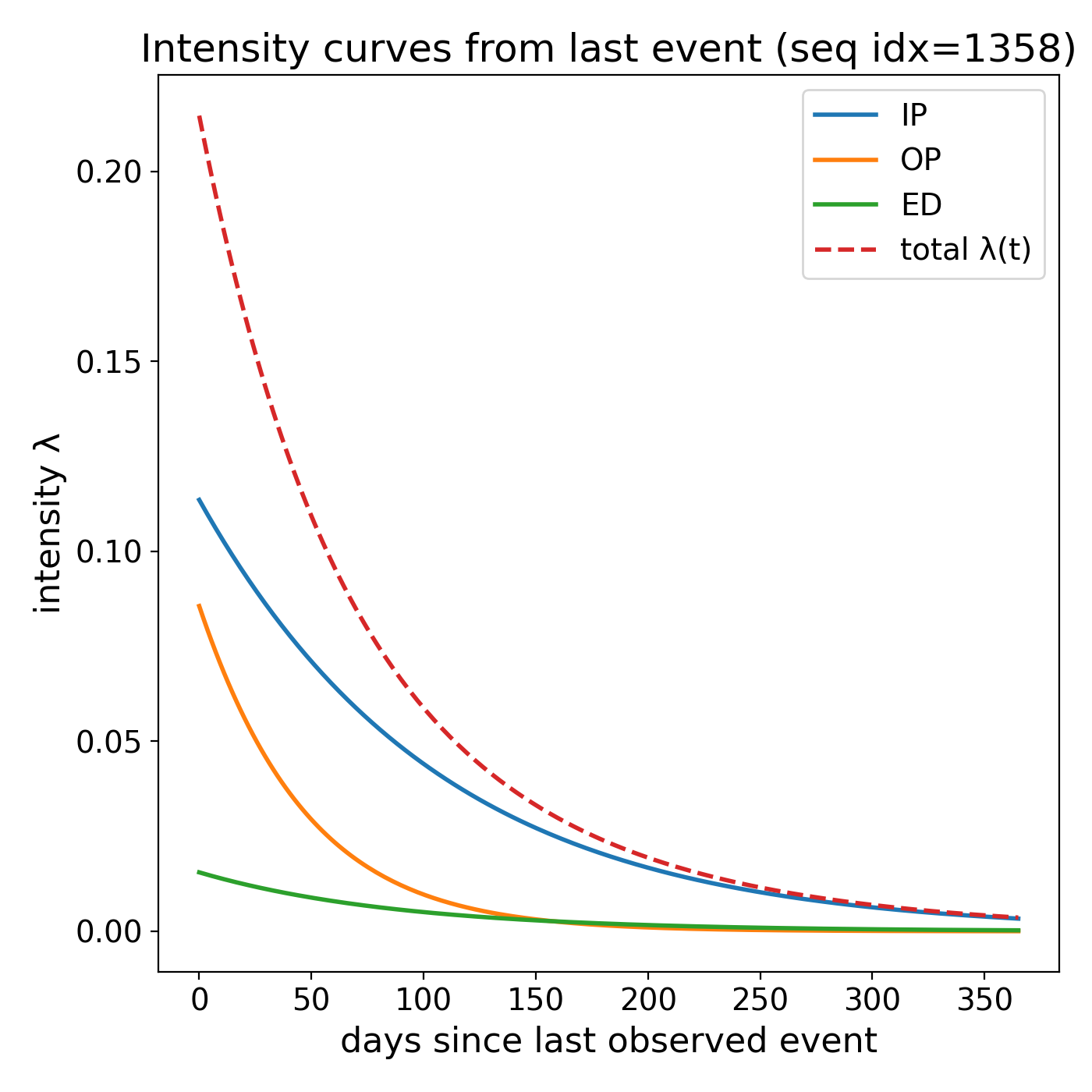}}
\hspace{0.02\textwidth}
\subfloat[Attention recency curve]{%
  \includegraphics[width=0.31\textwidth]{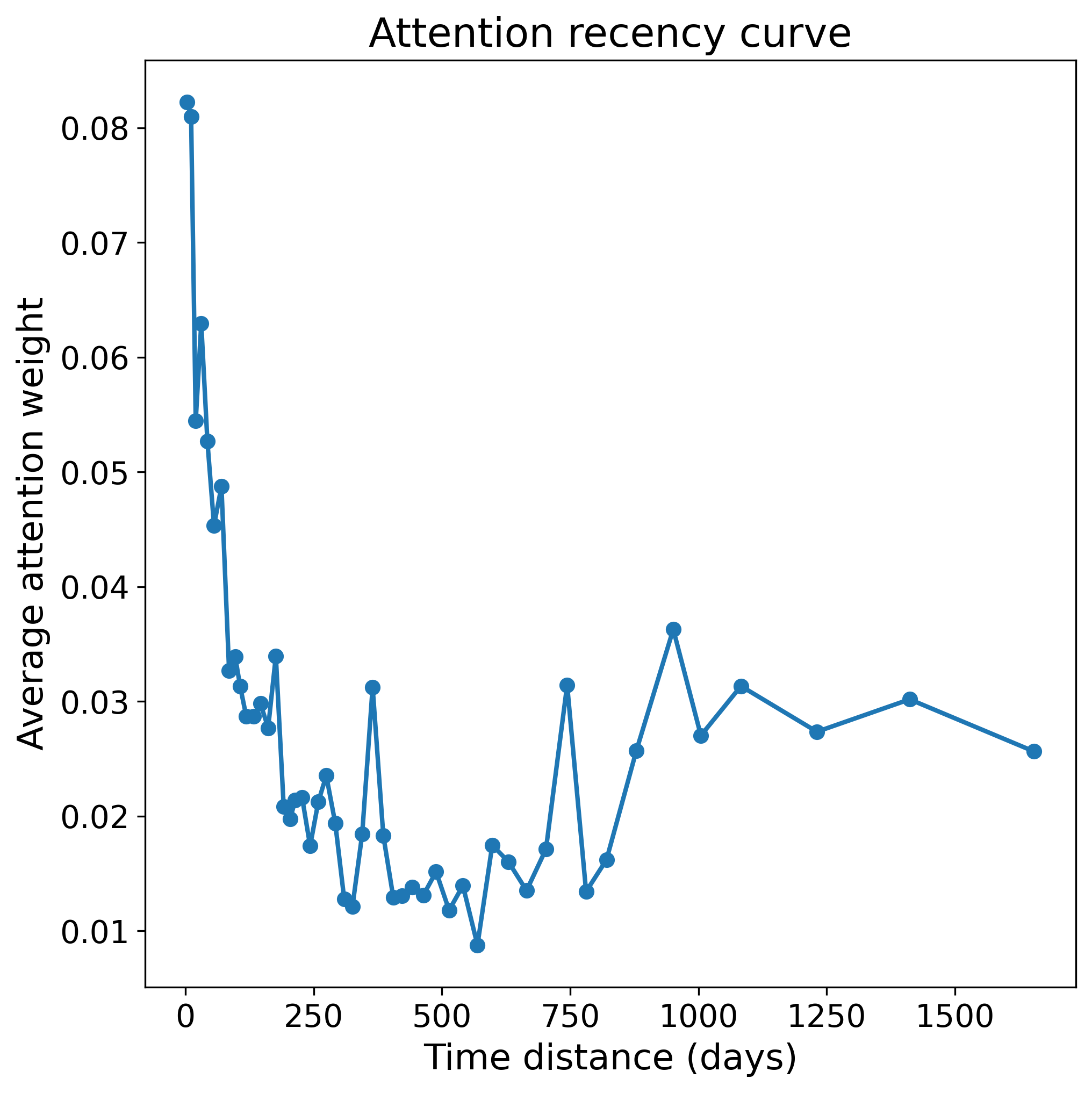}}
\hspace{0.02\textwidth}
\subfloat[Self-attention heatmap]{%
  \includegraphics[width=0.31\textwidth]{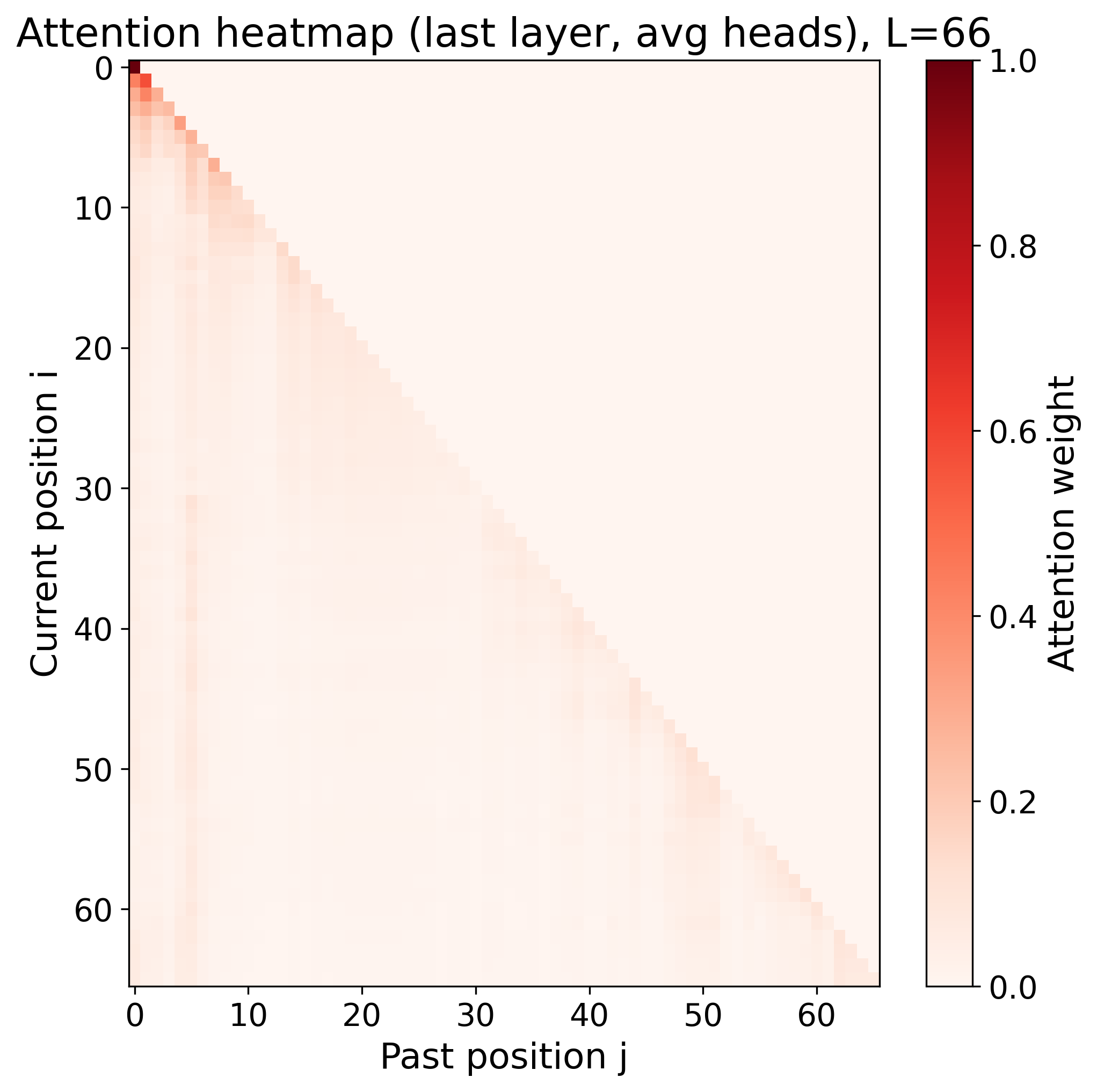}}
\caption{Interpretability analysis for an inpatient admission-dominant patient (Patient 1358).}
\label{fig:ip_interpretability}
\end{figure*}

\begin{figure*}[t]
\centering
\subfloat[Conditional intensity curves]{%
  \includegraphics[width=0.31\textwidth]{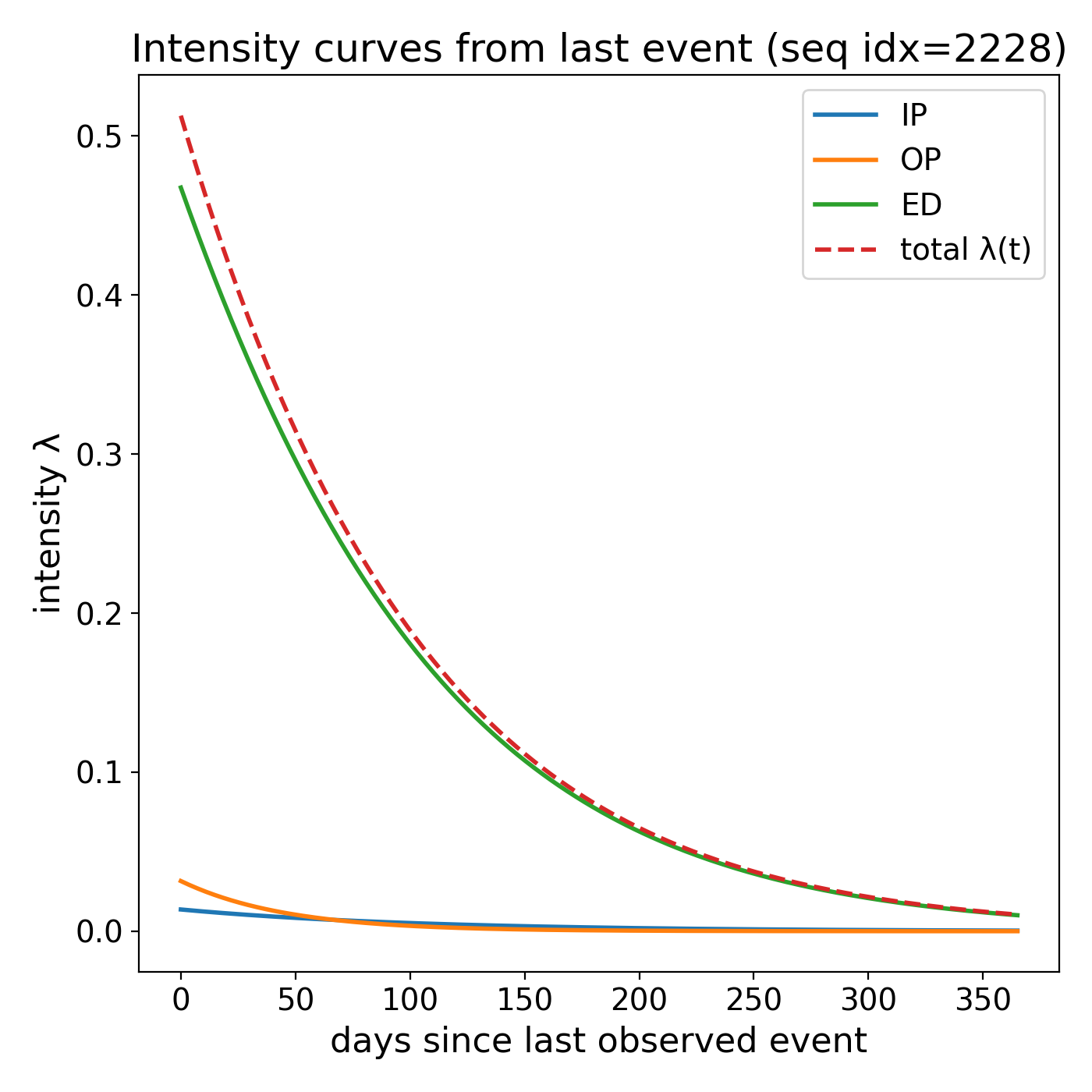}}
\hspace{0.02\textwidth}
\subfloat[Attention recency curve]{%
  \includegraphics[width=0.31\textwidth]{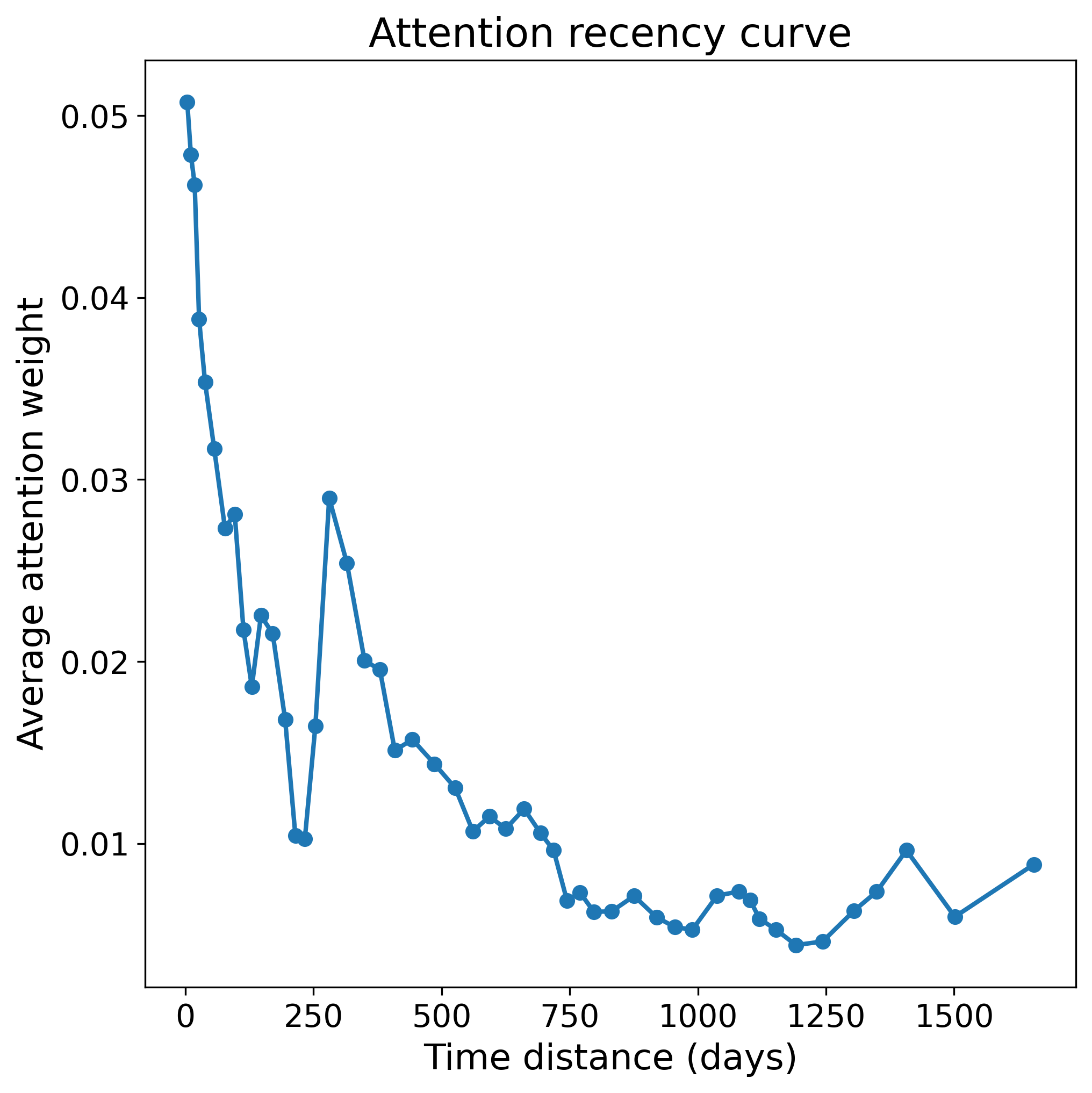}}
\hspace{0.02\textwidth}
\subfloat[Self-attention heatmap]{%
  \includegraphics[width=0.31\textwidth]{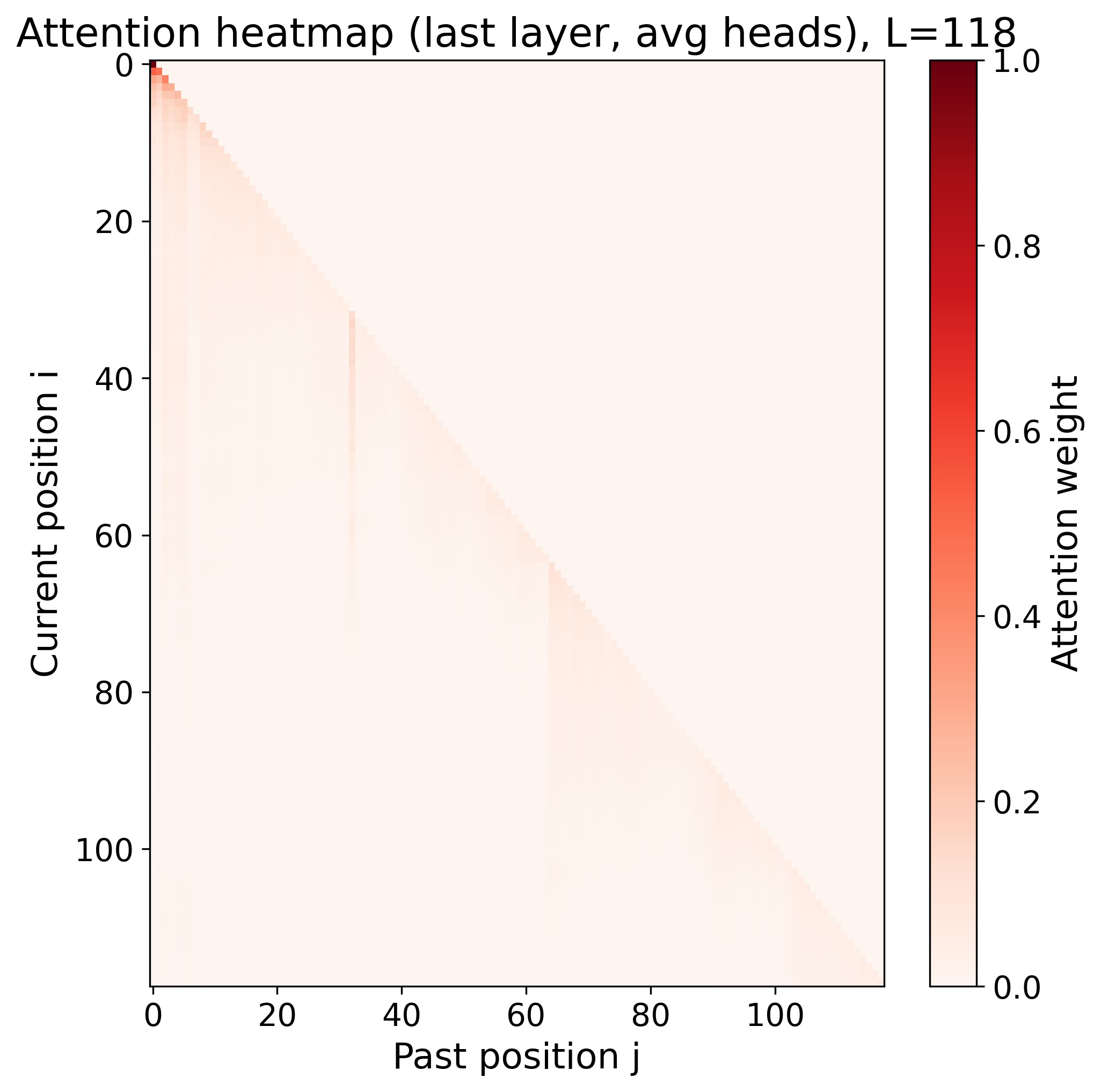}}
\caption{Interpretability analysis for an emergency department-dominant patient (Patient 2228).}
\label{fig:ed_interpretability}
\end{figure*}

\begin{figure*}[t]
\centering
\subfloat[Conditional intensity curves]{%
  \includegraphics[width=0.31\textwidth]{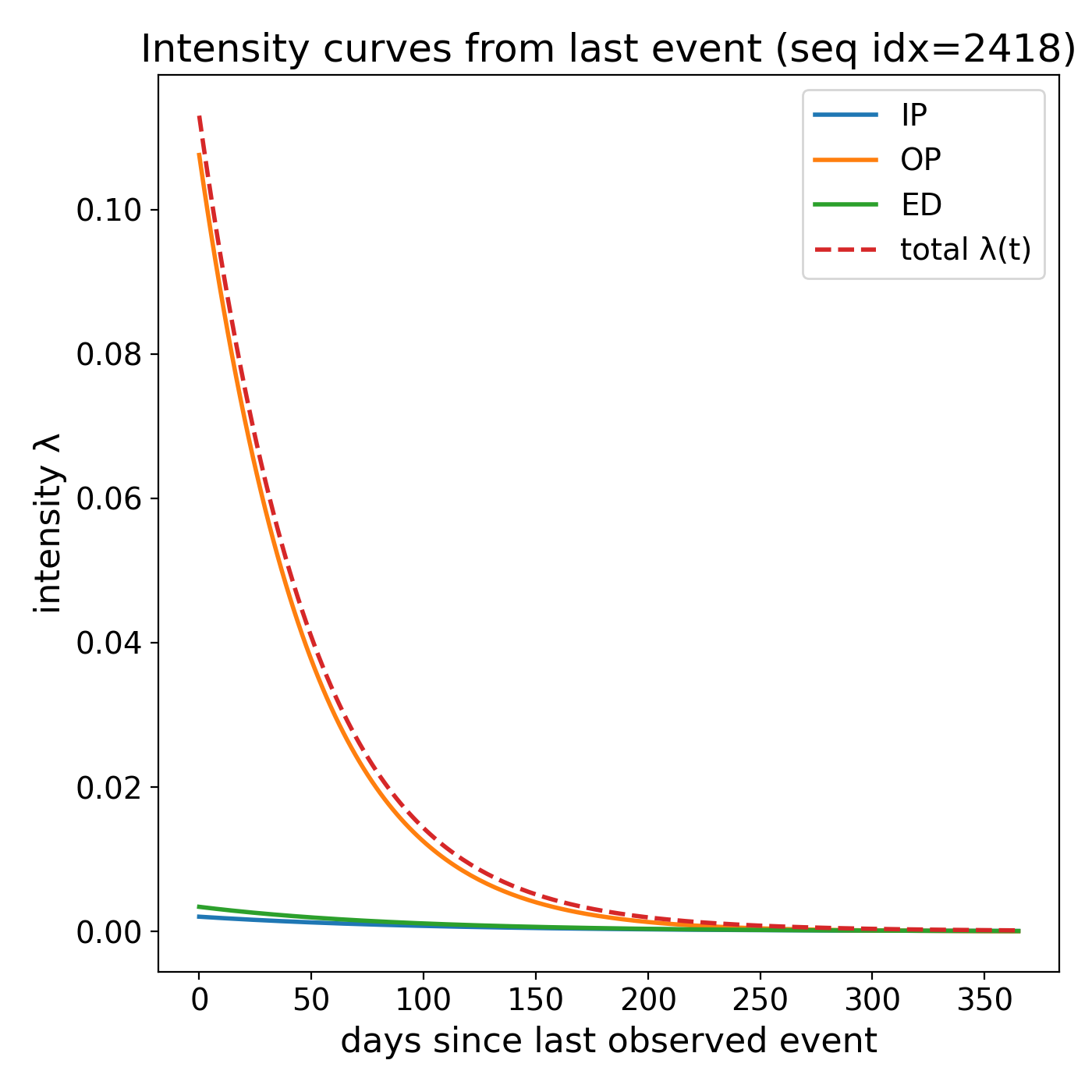}}
\hspace{0.02\textwidth}
\subfloat[Attention recency curve]{%
  \includegraphics[width=0.31\textwidth]{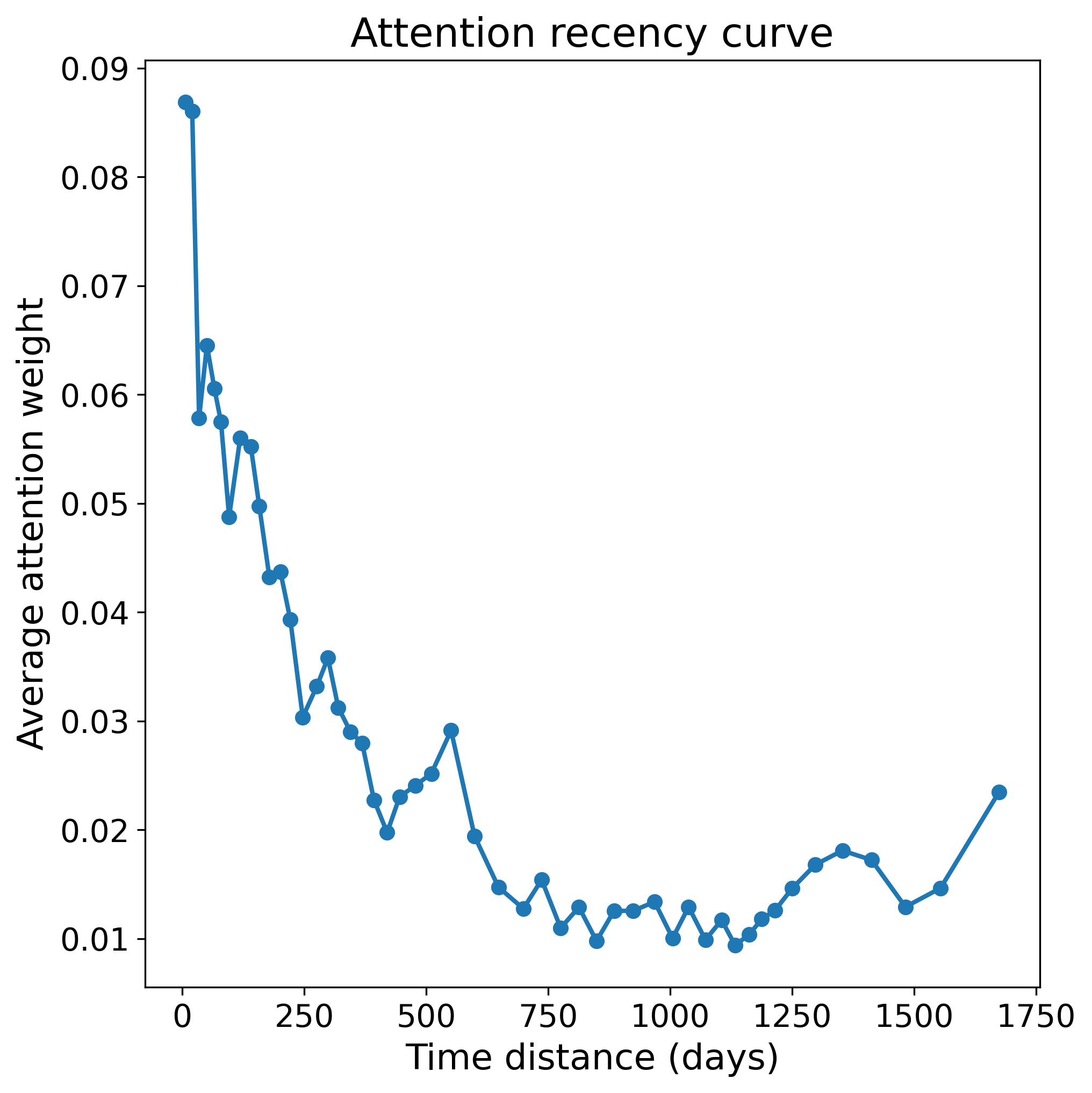}}
\hspace{0.02\textwidth}
\subfloat[Self-attention heatmap]{%
  \includegraphics[width=0.31\textwidth]{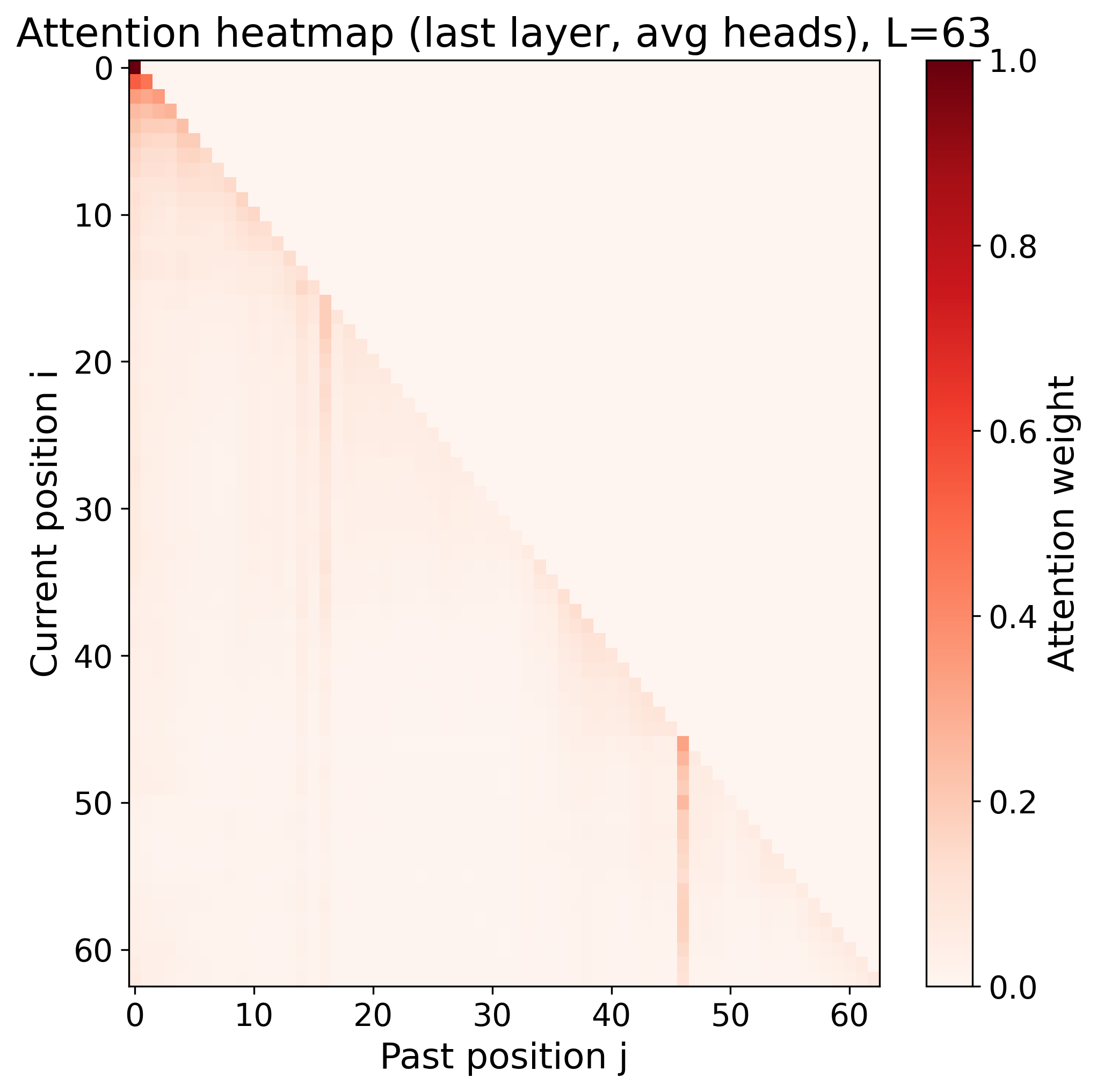}}
\caption{Interpretability analysis for an outpatient visit-dominant patient (Patient 2418).}
\label{fig:op_interpretability}
\end{figure*}

\section{Experimental Results}

\subsection{Experimental Setup}

All experiments are conducted on preprocessed patient event sequences using fixed training and test splits. The proposed Transformer Hawkes Process model is trained for up to 150 epochs with a batch size of 16. Optimization is performed using AdamW with learning rate scheduling and gradient clipping to ensure stable training. All reported results are obtained from the best-performing checkpoint on the test set.

\subsection{Data Description}

We evaluate our model using a real-world longitudinal healthcare utilization dataset derived from electronic health records (EHRs). The dataset comprises irregularly time-stamped event sequences representing patient healthcare encounters over time.

The data span six years (2019–2024) and include records from approximately two million patients. The cohort consists of high-need, high-cost patients enrolled in the Personalized Cross-Sector Transitional Care Management (PC-TCM) program, an initiative supported by the Agency for Healthcare Research and Quality (AHRQ). The dataset captures longitudinal inpatient, outpatient, and emergency department utilization patterns across the state of Western New York.

Each patient trajectory is modeled as an ordered sequence of clinical events, with each event associated with a timestamp denoting its occurrence time. All patient records were de-identified prior to analysis, and the study was conducted on fully anonymized data in compliance with applicable data use agreements and privacy regulations.

To ensure sufficient longitudinal information for temporal point process modeling, we further restricted the cohort to patients with at least 20 recorded clinical events. This filtering step resulted in a final study cohort of 42,184 patient trajectories. Patients with fewer events were excluded due to insufficient history to support reliable estimation of event-driven temporal dependencies. We additionally excluded the deceased patients during the period. This criterion was applied uniformly across all patients and was independent of event type or clinical outcome.

The dataset includes three types of clinical events:
\begin{itemize}
    \item \textbf{Inpatient admissions (IP)}: 51,988 events
    \item \textbf{Outpatient department visits (OP)}: 1,822,901 events
    \item \textbf{Emergency department encounters (ED)}: 142,850 events
\end{itemize}

In total, the dataset comprises 2,017,739 clinical events.

The sequence characteristics of the data are as follows:
\begin{enumerate}
    \item Variable-length patient trajectories, reflecting heterogeneous healthcare utilization patterns.
    \item Highly irregular inter-event times, with no fixed temporal spacing between consecutive events.
    \item Severe class imbalance, with outpatient visits dominating the dataset, followed by emergency encounters and relatively sparse inpatient admissions -- this imbalance poses a significant challenge for event-type prediction, as rare but clinically important events are easily overshadowed by dominant outpatient encounters.
\end{enumerate}

\subsection{Prediction Task}

Given a patient’s historical sequence of clinical events up to the current time, the prediction task is twofold:
\begin{enumerate}
    \item \textbf{Event-type prediction}: predicting the type of the next clinical event (IP, OP, or ED).
    \item \textbf{Time-to-event prediction}: estimating the time until the next event occurs, measured in days since the most recent event.
\end{enumerate}

This task is naturally formulated within a temporal point process framework, where both the event type and the event time of occurrence are modeled jointly.

\subsection{Evaluation Metrics}

Model performance is evaluated on both event-type prediction and time-to-event estimation tasks. 
For event-type prediction, we report the macro-averaged F1 score along with per-class F1 scores for inpatient (IP), emergency department (ED), and outpatient (OP) events. 
Macro-F1 is emphasized to account for the severe class imbalance inherent in healthcare utilization data and to ensure balanced assessment across both frequent and rare event types.

For time-to-event prediction, performance is evaluated using median absolute error (MedAE), measured in days. 
Healthcare inter-event times are highly skewed and heavy-tailed, and metrics such as RMSE are disproportionately influenced by a small number of extreme gaps. 
MedAE provides a robust estimate of typical temporal prediction error and is therefore more representative of real-world clinical performance. 
All metrics are computed on the held-out test set and, where relevant, stratified by event type.

\subsection{Quantitative Results}

Table~\ref{tab:merged_performance} summarizes the quantitative performance of the proposed Transformer Hawkes Process (THP) in comparison with GLM and LSTM-based Hawkes baselines. 
To ensure a fair comparison under severe class imbalance, both the LSTM-based Hawkes model and THP are trained using the same weighted cross-entropy loss, with class weights defined by the inverse square root of event frequencies. 
This controlled setup isolates the effect of architectural design from that of imbalance-aware optimization.

Across models, performance is strongly influenced by the dominance of outpatient (OP) events; however, substantial differences emerge in the ability to recognize rare but clinically critical inpatient (IP) and emergency department (ED) events. 
The GLM baseline largely fails to identify these minority classes, yielding near-zero F1 scores due to its strong bias toward the majority OP class. 
Incorporating temporal dependencies through an LSTM-based Hawkes formulation improves minority-class recognition to some extent, but performance remains skewed toward OP events despite imbalance-aware training.

In contrast, the proposed Transformer Hawkes Process (THP) achieves the highest macro-averaged F1 score, driven by consistent improvements in both IP and ED prediction while preserving strong OP performance. 
These gains arise even though both neural baselines optimize the same weighted cross-entropy loss, indicating that architectural differences—rather than loss design alone—play a decisive role. 
The attention-based history encoding in THP enables more effective utilization of imbalance-aware supervision by selectively emphasizing clinically relevant historical events, leading to improved modeling of heterogeneous and rare-event dynamics.

For time-to-event prediction, THP consistently achieves lower MedAE compared to both GLM and LSTM-based Hawkes baselines, indicating superior typical-case accuracy in estimating event timing. 
This robustness is particularly important in clinical settings, where accurate prediction of common time-to-event patterns is often more actionable than minimizing worst-case deviations driven by rare extremes.

Overall, these results demonstrate that while imbalance-aware loss functions are necessary for learning under realistic healthcare data distributions, model architecture critically determines how effectively such supervision is translated into clinically meaningful predictions. 
Although absolute improvements in macro-F1 and MedAE may appear modest, they are practically significant in highly imbalanced healthcare settings, where even small gains in predicting rare but critical events such as IP admissions and ED visits can have substantial downstream impact on clinical decision-making and resource planning.

\subsection{Ablation Study: Effect of Weighted Event-Type Loss}
\begin{table*}[t]
\centering
\caption{Ablation on Class-Imbalance Handling in THP}
\label{tab:ablation}
\begin{tabular}{lccccc}
\toprule
Model & Macro-F1 & F1(IP) & F1(ED) & F1(OP) & MedAE \\
\midrule
THP (w/o weighted CE) & 0.368 & 0.001 & 0.147 &0.955 & 13.58 \\
THP (weighted CE) & \textbf{0.480} & \textbf{0.162} & \textbf{0.334} & 0.942 & 13.00 \\
\bottomrule
\end{tabular}
\end{table*}

To assess the role of imbalance-aware supervision, we conduct an ablation in which the weighted cross-entropy loss for event-type prediction is removed from the Transformer Hawkes Process (THP). As shown in Table~\ref{tab:ablation}, removing class weighting preserves strong performance on the dominant outpatient (OP) class but leads to a near-complete collapse in inpatient (IP) recognition, with F1 scores approaching zero.

While emergency department (ED) prediction remains non-zero, performance is substantially degraded compared to the weighted variant. These results demonstrate that architectural advances alone are insufficient to address extreme class imbalance, and that imbalance-aware loss functions are critical for achieving clinically meaningful sensitivity to rare but high-impact events.

\subsection{Model Interpretability and Case Studies}

To assess the clinical interpretability of the proposed Transformer Hawkes Process (THP), we analyze three complementary signals derived from the model: (i) event-type--specific conditional intensity curves, (ii) attention recency profiles, and (iii) self-attention heatmaps. Together, these visualizations illustrate how THP integrates recent clinical context, long-term patient history, and event-type--specific risk dynamics when estimating future healthcare utilization.

We present representative case studies corresponding to inpatient-dominant, emergency department--dominant, and outpatient-dominant utilization patterns.

\subsubsection{Inpatient-Dominant Patient Trajectory}

\textbf{Patient 1358} exhibits an inpatient-dominant utilization profile with 66 total events, including 44 inpatient (IP) visits (66.7\%), 21 outpatient (OP) visits (31.8\%), and 1 emergency department (ED) visit (1.5\%).

\paragraph{Conditional Intensity Dynamics}
The event-type--specific conditional intensity curves (Fig.~\ref{fig:ip_interpretability}a) indicate that immediately following the most recent hospital encounter, the predicted risk of inpatient admission is substantially higher than that of outpatient or emergency visits. Outpatient risk remains moderate, while emergency risk is negligible, closely reflecting the patient’s historical utilization pattern. As time progresses, the intensities of all event types decay; however, inpatient risk remains elevated for an extended duration, suggesting sustained clinical vulnerability rather than a transient acute episode.

\paragraph{Attention-Based Interpretability}
The attention recency curve (Fig.~\ref{fig:ip_interpretability}b) shows that the model assigns the highest importance to recent events, with attention weights decaying rapidly as temporal distance increases. Importantly, non-zero attention persists for distant events, indicating that the model retains awareness of long-term utilization patterns such as recurrent admissions. The self-attention heatmap (Fig.~\ref{fig:ip_interpretability}c) exhibits a triangular causal structure, with strong concentration near the diagonal and selective vertical bands corresponding to salient historical inpatient encounters that influence multiple future events. While attention weights do not imply causal relationships, they provide a useful attribution signal indicating which historical events the model emphasizes when forming future predictions.

\paragraph{Clinical Interpretation}
This trajectory is clinically intuitive: patients with frequent inpatient admissions face the highest readmission risk shortly after discharge, with risk gradually tapering over time. The persistence of non-zero inpatient intensity reflects chronic disease burden or unresolved medical complexity. Overall, THP captures both short-term readmission risk and long-term vulnerability in a clinically meaningful manner.

\subsubsection{Emergency Department--Dominant Patient Trajectory}

\textbf{Patient 2228} represents an ED-dominant utilization pattern with 118 total events, including 113 ED visits (95.8\%), 4 OP visits (3.4\%), and 1 IP admission (0.8\%).

\paragraph{Conditional Intensity Dynamics}
As shown in Fig.~\ref{fig:ed_interpretability}a, the predicted ED intensity immediately after the most recent encounter is overwhelmingly dominant, far exceeding both inpatient and outpatient risks. Over time, ED risk declines gradually but remains non-zero for extended periods, indicating persistent instability or unmet care needs. In contrast, inpatient and outpatient intensities remain negligible throughout.

\paragraph{Attention-Based Interpretability}
The attention recency curve (Fig.~\ref{fig:ed_interpretability}b) demonstrates a strong emphasis on very recent events, with attention decaying sharply as temporal distance increases. The self-attention heatmap (Fig.~\ref{fig:ed_interpretability}c) shows minimal influence from distant history, indicating that ED utilization is primarily driven by recent encounters rather than long-range structured care patterns.

\paragraph{Clinical Interpretation}
This profile is characteristic of frequent ED utilizers, often associated with limited outpatient follow-up or chronic disease instability. THP correctly identifies ED revisits as the dominant and persistent risk, highlighting opportunities for ED diversion, improved discharge planning, and targeted care coordination.

\subsubsection{Outpatient-Dominant Patient Trajectory}

\textbf{Patient 2418} exhibits an outpatient-dominant utilization pattern with 63 total events, including 60 OP visits (95.2\%), 2 IP admissions (3.2\%), and 1 ED visit (1.6\%).

\paragraph{Conditional Intensity Dynamics}
The predicted intensity curves (Fig.~\ref{fig:op_interpretability}a) show a strong immediate likelihood of outpatient follow-up after the most recent visit, with minimal risk for inpatient or emergency events. Over time, outpatient risk decreases steadily, consistent with routine scheduled care cycles.

\paragraph{Attention-Based Interpretability}
Attention recency profiles (Fig.~\ref{fig:op_interpretability}b) indicate that recent outpatient visits exert the strongest influence on future predictions, while older visits retain a small but persistent effect. The self-attention heatmap (Fig.~\ref{fig:op_interpretability}c) exhibits diffuse, low-intensity attention without dominant historical events, reflecting a stable and predictable utilization pattern.

\paragraph{Clinical Interpretation}
This trajectory corresponds to a low-acuity patient managed primarily through ambulatory care. THP accurately differentiates stable outpatient-managed patients from high-acuity inpatient or ED-dominant trajectories, reinforcing its ability to capture heterogeneous healthcare utilization patterns.

\section{Conclusion}

This work studied the problem of modeling patient healthcare utilization as an irregular, multi-type event sequence and investigated the use of Transformer Hawkes Processes for continuous-time prediction under severe class imbalance. By representing patient trajectories as time-stamped inpatient, outpatient, and emergency department events, the proposed framework jointly models event-type interactions and temporal dynamics without relying on discretized time or fixed observation windows.

Experimental results on real-world healthcare utilization data demonstrate that the proposed approach effectively handles highly imbalanced care trajectories, yielding improved recognition of clinically important but infrequent events while preserving strong performance on dominant outpatient patterns. Beyond predictive performance, a key strength of the framework lies in its interpretability. Through conditional intensity curves, attention recency profiles, and self-attention visualizations, the model offers insight into how recent encounters and longer-term history contribute to future utilization risk. The presented case studies illustrate clinically intuitive patterns, including elevated short-term risk following inpatient discharge, persistent emergency department utilization driven by recent instability, and stable outpatient follow-up trajectories.

Overall, this work highlights the practical advantages of combining the principled structure of temporal point processes with the representational flexibility of transformer architectures. By enabling individualized, interpretable modeling of imbalanced care trajectories in continuous time, the proposed approach provides a robust foundation for temporal risk prediction and decision support in the real-world healthcare settings.

Although this work focuses on healthcare utilization data derived from electronic health records (EHRs), the proposed methodology is inherently domain-agnostic and can be applied to a wide range of event-driven datasets. The framework models data as sequences of irregularly time-stamped events with associated types, a structure that naturally arises in many real-world domains beyond healthcare.

The Transformer Hawkes Process formulation operates directly in continuous time and does not rely on domain-specific assumptions about event semantics. As a result, it can be readily extended to applications such as user activity modeling in online systems, financial transaction streams, system logs, and other temporal event sequences where both the timing and type of events are critical.

Furthermore, the imbalance-aware training strategy introduced in this work is not specific to healthcare and is applicable to any setting with skewed event-type distributions. By decoupling temporal modeling from domain-specific feature engineering, the proposed approach provides a general and flexible framework for modeling complex, heterogeneous, and imbalanced event sequences across diverse application areas.

\section{Future Work}

Future work will extend this framework to incorporate richer patient-level context, including demographic information, longitudinal disease history, and socioeconomic or financial indicators, enabling more comprehensive modeling of individualized care trajectories beyond event timing alone. Such extensions may further enhance both predictive performance and clinical relevance.

Another important direction is to strengthen interpretability, as explainable artificial intelligence remains critical for trust and adoption in healthcare. While attention-based explanations provide useful attribution signals, future work may explore complementary explanation strategies that better characterize uncertainty and model reliability.

Finally, improving performance on rare but clinically significant event types remains an open challenge. Although this work addresses severe class imbalance through cost-sensitive learning and achieves improved minority-class F1 scores, future research may investigate more advanced imbalance-aware strategies to further enhance prediction of critical outcomes without compromising overall model stability.

\section*{Acknowledgment}
This work was supported by AHRQ grant R01~HS028000. Computing support was provided by the Center for Computational Research at the University at Buffalo.


\bibliographystyle{IEEEtran}
\bibliography{references}

\end{document}